\documentclass[graybox]{svmult}

\usepackage{mathptmx}       % selects Times Roman as basic font
\usepackage{helvet}         % selects Helvetica as sans-serif font
\usepackage{courier}        % selects Courier as typewriter font
\usepackage{type1cm}        % activate if the above 3 fonts are
\usepackage{makeidx}         % allows index generation
\usepackage{graphicx,epsfig,color}
\usepackage{multicol}        % used for the two-column index
\usepackage[bottom]{footmisc}% places footnotes at page bottom

\usepackage{latexsym}
\usepackage{color}
\usepackage{subfig}
\usepackage{amssymb,amsfonts,amsmath} %for display mathmode 
\usepackage{bm} %\\boldsymbol command from package amsbsy (loaded by amsmath), but i have to attach this package then it will display correctly
\usepackage{subfig} % for subfigures in 2 columns
\usepackage{booktabs, dcolumn} % \midrule, \toprule

\newcommand{\x}{{\textbf{\emph{x}}}}

\newcommand{\w}{{\textbf{\emph{w}}}}

\newcolumntype{d}[1]{D{.}{.}{#1}} %tab
\makeindex             % used for the subject index
\begin{document}
\title*{Simplifying the minimax disparity model for
determining OWA weights in large-scale problems}
\titlerunning{Simplifying the minimax disparity model}

%
\author{Thuy Hong Nguyen}
\institute{ Thuy Hong Nguyen \at Department of Industrial Engineering, University of
Trento,\\ 
Via Sommarive 9, I-38123 Povo (TN), Italy, \email{hongthuy.nguyen@unitn.it}
}
\maketitle
\abstract{In the context of multicriteria decision making, the ordered weighted averaging (OWA) functions play a crucial role in aggregating multiple criteria evaluations into an overall assessment supporting the decision makers' choice. Determining OWA weights, therefore, is an essential part of this process. Available methods for determining OWA weights, however, often require heavy computational loads in real-life large-scale optimization problems. In this paper, we propose a new approach to simplify the well-known minimax disparity model for determining OWA weights. For this purpose, we use to the binomial decomposition framework in which natural constraints can be imposed on the level of complexity of the weight distribution. The original problem of determining  OWA weights is thereby transformed into a smaller scale optimization problem, formulated in terms of the coefficients in the binomial decomposition. Our preliminary results show that a small set of these coefficients can encode for an appropriate full-dimensional set of OWA weights.
 }
%\linebreak 
%\textbf{Keywords} Ordered weighted averaging, OWA weights determination, binomial decomposition framework, k-order additivity, large-scale optimization problems.}

\section{Introduction}
\label{sec:1}
%In many disciplines, decision makers usually deal with  problems involving the process of aggregating  multi-arguments and producing the overall objective functions associated with the application context. 
%The ordered weighted averaging function (OWA) introduced by Yager \cite{Yager-1988} is one of the most important aggregation functions in decision making theory. 
%
% One of the main motivations behind selecting the OWA functions  for aggregation of multi-criteria is its flexibility to provide a general class of parameterized aggregation functions bounded between the \emph{min} and the \emph{max} functions, which is controlled by their weight distributions. 
%revision
In many disciplines, decision makers have to deal with problems involving the aggregation of  multicriteria evaluations and producing overall assessments within the application context. The ordered weighted averaging function
(OWA) introduced by Yager \cite{Yager-1988} is a fundamental aggregation function
in decision making theory. One of the main motivations behind selecting the OWA
functions for multicriteria aggregation is their flexibility in providing a general class
of weighted aggregation functions bounded by the \emph{min} and the \emph{max} functions.

%%OWA's application 
% Since then many applications in various research areas have investigated the usage of the OWA functions, such as,  fuzzy logic controllers (Yager \cite{Yager-1991,Yager-Filev-1992}), vision systems (L\'{o}pez et al. \cite{Lopez-Alvarez-Millan-Puig-Torra-1992}), multi-criteria decision making (Yager \cite{Yager-1988, Yager-1993, Yager-1992}, Marichal \cite{Marichal-1998}), expert systems (Carlsson \cite{Carlsson-Fuller-Fuller-1997}, O'Hagan \cite{OHagan-1988}, Bortot et al. \cite{Bortot-MarquesPereira-Thuy-2017}), neural networks (Yager \cite{Yager-1987,Yager-1992-IJIS,Yager-1992-IJNN}), see also Yager and Kacprzyk \cite{Yager-Kacprzyk-1997}, and Yager et al. \cite{Yager-Kacprzyk-Beliakov-2011}.
% %
%Therefore, the determination of the OWA weights becomes a very important study of applying OWA functions in decision making theory.
 
The determination of appropriate OWA weights is thus a very important object of study when applying OWA functions in the context of decision making. Among the various methods for obtaining OWA weights that can be found in the literature,  namely \cite{Xu-2005,Xu-2003,Yager-1993}, and more recently \cite{Beliakov-BustinceSola-Calvo-2016, Carlsson-Fuller-2018} and \cite{Liu-2011}, we can distinguish the following methodological categories: a) methods based on learning OWA weights from data, b) methods based on weight-generating functions, and c) methods based on the  characteristic measures of \emph{orness}  and \emph{disparity}.

%
%The Miness or Maxness level can be measured by the "andness" or "orness" indicators equivalently. 

%Among a large number of methods introduced for obtaining OWA weights mentioned in \cite{Xu-2005,Xu-2003,Yager-1993}, see also more recently  \cite{Beliakov-BustinceSola-Calvo-2016, Carlsson-Fuller-2018} and \cite{Liu-2011}, we  categorize them into three primary groups based on their approaches: a) learning OWA weights from data, b) methods based on weight generating functions, and c) methods based on OWA characterizing measures : dispersion and orness. 
%
%%
%In this paper we mainly focus on the problems of determining a special class of OWA functions based on the dispersion measure under a given level of \emph{orness} since these two characterizing  measures can establish OWA weights associated with different attitudinal modes of decision makers, see  \cite{Mezei-Brunelli-2017, Yager-1988}. Depending on the weighting structure of the OWA functions, the associated OWA functions reflect different preferences of decision makers, from the optimistic to the pessimistic attitudes. 
%%
%The level of attitudinal characters of decision makers is measured by their \emph{orness}, which is bounded in the unit interval . The maximum (minimum) orness value is attained when decision makers are purely optimistic (purely pessimistic).
%In other hands, the \emph{dispersion} (or \emph{entropy}) measure evaluates the level of importance of input arguments  are taken into account in the aggregation process. 
In this paper we  focus on the problem of determining a special class of OWA functions based on the disparity measure under a given level of orness, since these two  measures can lead to OWA weights associated with different attitudinal characters of
decision makers, see \cite{Mezei-Brunelli-2017, Yager-1988}. Depending on their weighting structure, OWA functions reflect different preferences of decision makers, from the optimistic to the pessimistic attitude. The attitudinal character of decision makers is measured by their orness, which takes values in the unit interval. The maximum (minimum) orness value is attained when decision makers
are purely optimistic (purely pessimistic).
% In other hands, the \emph{dispersion} (or \emph{entropy}) measure evaluates the level of importance with which input arguments are taken into account in the aggregation process.
 On the other hand, the disparity  evaluates the non-uniformity of the OWA weights. In the case of the arithmetic mean the disparity takes its minimal value.

In the literature several methods have been introduced to obtain the optimal weights by using the disparity measure. After the pioneering work of O'Hagan \cite{OHagan-1988} on the maximal entropy method
and the variance-based methods of Yager \cite{Yager-IJUFKBS-1996} and Fuller and Majlender \cite{Fuller-Majlender-2003}, Wang
and Parkan \cite{Wang-Parkan-2005} proposed the minimax disparity method in which the objective is to minimize the maximum absolute difference between two adjacent weights. 
%One of the advantage of the minimax disparity approach is that it uses simple linear programming 
%in order to obtain the optimal OWA weights.
Liu \cite{Liu-2007}  proved the
equivalence of the solutions of the minimum variance approach suggested by Fuller
and Majlender \cite{Fuller-Majlender-2003} and the minimax disparity model proposed by Wang and Parkan
\cite{Wang-Parkan-2005} under a given level of orness. Extensions of disparity-based
models for determining OWA weights are presented in \cite{Amin-Emrouznejad-2006, Emrouznejad-Amin-2010, Gong-Dai-Hu-2016, Sang-Liu-2014,Wang-Luo-Hua-2007}. 
In this paper, we focus on the minimax disparity model since it  has recently received a great deal of interest in the literature and is easy to solve due to its simple linear programming formulations.

The usual academic instances of the minimax disparity model focus on solving problems with small dimensions ($n = 3, 4, 5, 6$). However, in applied operational research, optimization problems are often much more complex and require a heavy computational demand when there 
are hundreds or thousands of variables. In order to overcome the complexity of high-dimensional problems, we consider the binomial decomposition framework, proposed by Calvo and De Baets \cite{Calvo-DeBaets-1998}, see also Bortot and Marques Pereira \cite{Bortot-MarquesPereira-2014} and Bortot et al. \cite{Bortot-Fedrizzi-MarquesPereira-Thuy-2017, Bortot-MarquesPereira-Thuy-2017}, which refers to the k-additive framework introduced by Grabisch \cite{Grabisch-1997b, Grabisch-1997c,Grabisch-1997a}. This framework allows us to transform the original problem, expressed directly in terms of the OWA weights, into a problem in which the weights are substituted by a new set of coefficients. In this transformed representation, we can consider only a reduced number of these coefficients, associated
with the first k-additive levels of the OWA function, and we can set the remaining coefficients to zero. In this way the computational demand in high-dimensional problems is significantly reduced. 

%In this paper, we choose the minimax disparity model since it has recently received a great interest in the literature and it is easy to solve due to its simple linear programming formulations.

%The paper is organized as follows. 
%In Sect. \ref{sec:2}  we briefly review the  ordered weighted averaging functions and their representation in the binomial decomposition framework. 
%%
%Section \ref{sec:3} reviews the recent development of the disparity-based models for determining OWA weights. 
%%
%In Sect. \ref{sec:4} we recall the minimax disparity model and reformulate it in terms of coefficients alpha and solve them with the  dimension $n=5$ which is also used for the test in the original problems.
%%In this framework, we underline that the linear relationship between OWA weights and the coefficients is one-to-one correspondence. 
%%Then we reformulated the original  disparity-based  models for determining OWA weights in terms of coefficients alpha and solve them in the  dimensions $n=5$ which is tested in the original problems. 
%We further complicate our empirical test  with larger dimensions, for instance $n=10$.
%%, we can start with $n=10$ and then we develop our test model to any large dimensions.
%% The optimal alpha solutions are used to calculate their associated OWA weights.
%%  Next the result is  compared with the optimal OWA weights obtain directly in the original problems in small and large dimensions.
%Finally, Sect. \ref{sec:5} contains some conclusive remarks.

The remainder of this paper is organized as follows. In Sect. \ref{sec:2} we briefly review the ordered weighted averaging functions and their representation in the binomial decomposition
framework. Section \ref{sec:3} reviews the recent development of the minimax disparity model for determining OWA weights. In Sect. \ref{sec:4} we recall the minimax disparity model and reformulate it in terms of the coefficients  in the binomial decomposition framework. We illustrate our approach for dimension $n=10$. Finally, Sect. \ref{sec:5} contains some conclusive remarks.

%THUY ADD
%%%%%%%%%%%%%%%%%%%%%%%%%%%%%%%%%%%%%%%%%%%%
\section{OWA functions and the binomial decomposition framework}\label{sec:2}

In this section we consider a point  $\x \in \mathbb{R}^n$, with   $\,n \ge 2$. The increasing  reordering of the coordinates of $\x $ is denoted as $x_{(1)}
\leq \dots \leq x_{(n)}$ . We now introduce the definition of the ordered weighted averaging function and its characterizing measures.

\begin{definition}\label{OWAdefinition}
An \emph{Ordered Weighted Averaging
(OWA) function} of dimension $n$ is an averaging
function $A:\mathbb{R}^n \longrightarrow \mathbb{R}$ with an associated weighting vector $\w =(w_1,\dots,w_n)\in[0,1]^n\!$,
such that $\sum_{i=1}^nw_i$ $=1$
and

%Given a weighting vector $\w =(w_1,\dots,w_n)\in[0,1]^n\!$,
%with $\sum_{i=1}^nw_i$ $=1$, the \emph{Ordered Weighted Averaging
%(OWA) function} associated with $\w $ is the averaging
%function $A:\mathbb{R}^n \longrightarrow \mathbb{R}$ defined as
%
\begin{equation}
A(\x ) = \sum_{i=1}^nw_i \, x_{(i)}.
\end{equation}
\end{definition}

Different OWA functions are classified by their weighting vectors. The OWA weights are characterized by two measures called orness  and disparity. 
In the following part we review these two measures and their properties. 

\begin{definition}
 Consider an OWA function  with an associated weighting vector $\w =(w_1,\dots,w_n)\in[0,1]^n$ such that $\sum_{i=1}^nw_i$ $=1$, two characterizing measures called \emph{orness} and \emph{disparity} are defined as

\begin{equation}
{Orness}(\w) =  \frac{1}{n-1}\sum_{i=1}^{n} (i-1)w_i,   \hspace{0.8cm}
Disparity(\w) = \max\limits_{i\in \{1,\ldots,n-1\}}   |w_i - w_{i+1}| .
\end{equation}
\end{definition} 
Yager \cite{Yager-1988} introduced the orness measures  to evaluate the level of similarity between the OWA function and the \emph{or (maximum)} operator. On the other hand, the disparity measure, as proposed by Wang and Parkan \cite{Wang-Parkan-2005}, is defined as the maximum absolute difference between two adjacent weights. Its value shows how unequally multicriteria evaluations  are taken into account in the aggregation process. Both OWA characterizing measures are  bounded in the unit interval.
 Three special OWA weighting vectors are  $\w_*=(1,0,\ldots,0)$, $\w_A=(\frac{1}{n},\ldots,\frac{1}{n})$ and $\w^*=(0,\ldots,0,1)$.
 For these vectors we have the orness equal to $0, 0.5, 1$ and disparity equal to $1, 0, 1$, respectively.

 In the following  we recall the binomial decomposition of the OWA functions as in Bortot and Marques Pereira\cite{Bortot-MarquesPereira-2014} and  Bortot et al. \cite{Bortot-Fedrizzi-MarquesPereira-Thuy-2017}.

\begin{definition}
The \emph{binomial OWA functions} $C_j:\mathbb{R}^n
\longrightarrow \mathbb{R}$, with $j=1,\ldots ,n$, are defined as
\begin {equation}\label{binomial.owa}
C_j(\x )= \sum_{i=1}^{n}\, w_{ji}\, x_{(i)} =
\sum_{i=1}^{n}\, \frac{\binom {n-i} {j-1}}{\binom {n} {j}}\, x_{(i)}
\qquad j=1,\ldots ,n
\end {equation}
where the binomial weights $w_{ji}$, $i,j=1,\ldots ,n,$ are null when
$i+j>n+1$, according to the usual convention that $\binom {p} {q}=0$
when $p<q$,  with $p,q=0,1,\ldots $. 
%In other words, the binomial OWA functions $C_j$ are rewritten shortly as $C_j(\x )= \sum_{i=1}^{n-j+1}\, w_{ji}\, x_{(i)}$. 
\end{definition}

%%%%%%%%%%%%%%%%%%%%%%%%%%%%%%test
%\begin{proposition}
%\label{OWA.decomposition.welfare.functions}
%%
%[Binomial decomposition] Any OWA function $A:\mathbb{R}^n \longrightarrow
%\mathbb{R}$ can be written uniquely as
%%
%\begin{equation}\label{OWA.decomposition.functions}
%A(\mathbb{\x})  = \alpha_1C_1(\x) +\alpha_2C_2(\x) + \cdots + \alpha_nC_n(\x)
%\end{equation}
%% 
%\end{proposition}
%%%%%%%%%%%%%%%%%%%%%%%%%%%%%%test

%
\begin{theorem}
%\begin{definition}
\label{OWA.decomposition.welfare.functions}
[Binomial decomposition] Any OWA function $A:\mathbb{R}^n \longrightarrow
\mathbb{R}$ can be written uniquely as
\begin{equation}\label{OWA.decomposition.functions}
A(\textbf{x})  = \alpha_1C_1(\textbf{x}) +\alpha_2C_2(\textbf{x}) + \cdots + \alpha_nC_n(\textbf{x})
\end{equation}
where the coefficients $\alpha_j$, $j=1,\ldots ,n,$ are subject to the
following conditions,
\begin{equation}\label{conditions.bsymm.alphak}
\alpha_1 =1-\sum_{j=2}^n\alpha_j \geq 0
\hspace{0.3cm} \text{ and } \hspace{0.3cm}
%
%\begin {equation}\label{conditions.msymm.alphak}
\sum_{j=2}^n \Big[1-n\frac{\binom{i-1}{j-1}}{\binom{n}{j}}\,
\Big]\alpha_j\leq 1 \qquad \qquad
 i=2,\dots,n
\end{equation}
%\end{definition}
\end{theorem}
The detail proof of Theorem \ref{OWA.decomposition.welfare.functions} is given in  Bortot and Marques Pereira \cite{Bortot-MarquesPereira-2014}.

The binomial decomposition (\ref{OWA.decomposition.functions}) expresses the linear combination between the OWA weights and the coefficients \emph{alpha} and can be written as the linear system

\begin{equation}\label{OWA.binomial.decomposition.linear.system.reduced}
%\qquad\Longleftrightarrow\qquad
\begin{aligned}[l]
\left\{
                \begin{array}{lcl}
                 w_1  &=&  w_{11}\alpha_1   + w_{21}\alpha_2   + \dots + w_{n-1,1}\alpha_{n-1}  +w_{n1}\alpha_n   \\
                 w_2  &=&  w_{12}\alpha_1   + w_{22}\alpha_2   + \dots + w_{n-1,2}\alpha_{n-1}  \\
                \makebox[2em]{\dotfill}\\
                w_{n-1}  &=&  w_{1,n-1}\alpha_1   + w_{2,n-1}\alpha_2   \\
                 w_n  &=&  w_{1n}\alpha_1   
                \end{array}
              \right.
\end{aligned}
\end{equation}
where the binomial weights  $w_{ji}=\frac{\binom {n-i}{j-1}}{\binom {n}{j}}, i, j = 1, \ldots, n,$ and the coefficients $\alpha_j, j=1, \ldots, n,$ are subjected to the conditions (\ref{conditions.bsymm.alphak}).

We notice that the coefficient matrix of the linear system is composed by the binomial weights $w_{ji}=\frac{\binom {n-i}{j-1}}{\binom {n}{j}}, i, j = 1, \ldots, n$, with the first $n-j+1$ weights being positive and non-linear decreasing, and the last $j-1$ weights equal to zero.  Hence there always exists a unique vector of  coefficients alpha satisfying the   linear system which is triangular and  whose coefficient matrix is full rank and invertible.

%%%%%%%%%%%%%%%%%%%%%%%%%%%%%%%%%%%%%%%%%%%%
%
%
%%%%%%%%%%%%%%%%%%%%%%%%%%%%%%%%%%%%%%%%%%%%
\section{The minimax disparity model for determining OWA weights}\label{sec:3}
The determination of appropriate OWA weights is  a very important study when applying OWA functions in the context of decision making. 
%One of important study of applying OWA functions in decision making theory is the study of determining the OWA weights. 
In this section, we briefly review the recent development of the specific class of OWA functions whose weights are determined by the minimax disparity methods. 

%\subsection{The entropy maximization}
%O'Hagan  \cite{OHagan-1988} was a pioneer introducing a maximum entropy approach to construct optimal weights at a given degree of orness. The optimization problem is the concave optimization problem with constraints
%\begin{eqnarray}\label{Model: Hagan.entropy.maximize}
%\text{Max.} && \hspace{0.3cm} -\sum_{i=1}^{n} {w_ilnw_i} \\
%\mbox{s.t.} \;  &&\text{orness}(\w) = \eta = \frac{1}{n-1}\sum_{i=1}^{n} (i-1)w_i  , \hspace{0.3cm} 0\leq \beta \leq 1\label{eq: predefined level of orness}\nonumber\\
%                &&\sum_{i=1}^{n} w_i =1 \label{eq: wi boundary}\nonumber\\
%                &&0\leq w_i \leq 1, \hspace{0.3cm} i, \ldots, n \nonumber\label{eq: wi boundary} %
%\end{eqnarray}

%\subsection{The minimax disparity approach }
%

In 2005 Wang and Parkan \cite{Wang-Parkan-2005} revisited  the maximum entropy method introduced by O'Hagan \cite{OHagan-1988} and proposed the minimax disparity procedure to determine the OWA weights in the convex optimization problem

%\begin{equation}
%\mathbb{\alpha} = (\alpha_1, \ldots, \alpha_n)\\
%\boldsymbol{\alpha} = (\alpha_1, \ldots, \alpha_n)\\
%\end{equation}
%\begin{eqnarray}
%%\text{Min.} \hspace{0.2} \left\lbrace  &=& \\
%
%\end{eqnarray}
\begin{eqnarray}\label{Model: Wand.Parkan.minimax.disparity.convex}
\text{Min.} && \hspace{0.3cm} \left\lbrace\max\limits_{i\in \{1,\ldots,n-1\}}   |w_i - w_{i+1}|\right\rbrace \label{eq:obj} \\
%\text{Min.} && \hspace{0.3cm} Disparity(\w) \label{eq:obj} \\
\mbox{s.t.} \;  &&\sum_{i=1}^{n} w_i =1 \label{eq: wi boundary}\nonumber, \hspace{0.5cm}
				\text{Orness}(\w) = \eta = \frac{1}{n-1}\sum_{i=1}^{n} (i-1)w_i  , \hspace{0.5cm}  \label{eq: predefined level of orness}\nonumber
                0\leq w_i \leq 1  \nonumber\label{eq: wi boundary} %
\end{eqnarray}
where $0\leq \eta \leq 1$ stands for the orness of the weighting vector.

The objective function is non-linear due to its formulation encompassing the absolute  difference between two adjacent weights. 
In order to overcome this non-linearity, the authors  introduced a new variable called $\delta=\max\limits_{i\in \{1,\ldots,n-1\}}   |w_i - w_{i+1}| \label{eq:obj}$ 
%
%\begin{eqnarray}  
%\delta=\max\limits_{i\in \{1,\ldots,n-1\}}   |w_i - w_{i+1}| \label{eq:obj} \\
%\end{eqnarray}
%
to denote the maximum absolute difference between two adjacent weights. This expression is  rewritten equivalently by two  inequations $
w_i-w_{i+1} - \delta\leq 0$ and $w_i-w_{i+1} +\delta\geq 0$.
The original optimization problem is thereby reformulated into the linear programming problem

\begin{eqnarray}\label{Model: Wand.Parkan.minimax.disparity.LP}
\text{Min.} && \delta \label{eq:Wang-Parkan-obj} \\
\mbox{s.t.} \;  &&\sum_{i=1}^{n} w_i =1 \label{eq: wi boundary}\nonumber, \hspace{1cm}
\text{Orness}(\w) = \eta = \frac{1}{n-1}\sum_{i=1}^{n} (i-1)w_i  , \hspace{0.5cm} \label{eq: predefined level of orness}\nonumber\\                
                &&w_i-w_{i+1} - \delta\leq 0 \label{eq: max 1}\nonumber, \hspace{0.5cm}
                w_i-w_{i+1} + \delta\geq 0 \label{eq: max 2}\nonumber,  \hspace{0.5cm}
                0\leq w_i \leq 1  \nonumber \label{eq: wi boundary} 
\end{eqnarray}
where $0\leq \eta \leq 1$ stands for the orness of the weighting vector.

The formulation  (\ref{Model: Wand.Parkan.minimax.disparity.LP}) is easy to solve in practice due to its linearity. 
Many researchers, therefore,  revisited this method and suggested numerous extensions  \cite{Amin-Emrouznejad-2006,Emrouznejad-Amin-2010,Gong-Dai-Hu-2016,Sang-Liu-2014,Wang-Luo-Hua-2007}. 
In particular Liu \cite{Liu-2007}  proved the
equivalence of the solutions of the minimax disparity model  and the minimum variance method suggested by Fuller
and Majlender \cite{Fuller-Majlender-2003}.

\section{The minimax disparity model and the binomial decomposition} \label{sec:4}
\noindent
In Sect. \ref{sec:3}, we have reviewed minimax disparity methods for determining OWA weights. The empirical results in those methods are carried out in the small dimensions $n=3,4,5,6$. In real-life problems, we usually encounter  large-scale optimization problems.  One of the challenges that clearly emerges  from the above methods in large-scale problems is that the optimization problems formulated directly in terms of the OWA weights require high computational resources.  Our objective is to improve the minimax disparity methods for determining  OWA weights in  high-dimensional problems. 

%Therefore, a new method of determining the OWA weights which can significantly reduce the problem complexity is important for future study.

In order to reduce the prohibitive complexity of the optimization problems, we recall the work of  Bortot and Marques Pereira \cite{Bortot-MarquesPereira-2014}. According to the authors, any OWA function can be decomposed uniquely into a linear combination of the binomial coefficients $\alpha_1, \ldots, \alpha_n$ and the binomial OWA functions $C_1, \ldots, C_n$, and can be written as $A(\x)  = \alpha_1C_1(\x) +\alpha_2C_2(\x) +
\ldots + \alpha_nC_n(\x)$, as described in Sect. \ref{sec:2}. 
The positive aspect  of the binomial decomposition framework is the k-additive levels in which a certain number of coefficients $\alpha_1,\ldots,\alpha_k$ are used in the process of  OWA weights determination while the remaining coefficients are set to zero. 
When the information of the coefficients alpha is available, we can reconstruct the OWA weights due to their one-to-one correspondence. In this way the computational demand in high-dimensional problems is significantly reduced.
%Moreover, due to the binomial weights $w_{ji}$ has null weight when $i+j>n+1$, the linear system can be reformulated as the upper triangle coefficent matrix of $w_{ji}$ times the binomial coefficients $\alpha_j$. The solution of the linear system, the OWA weight $\boldsymbol{w}$ in term of $\boldsymbol{\alpha}$, is unique and in one-to-one relationship with the coefficient $\alpha_j$.
%
%since the OWA weights are represented in terms of the coefficient $\alpha_j$.

%For the empirical test, we can select one of disparity-based methods. At the preliminary phase of this paper, we firstly apply our proposed method to the 
%minimax disparity model introduced by  Wang \& Parkan  \cite{Wang-Parkan-2005} since this model recently continues receiving a significant number of reviews and studies in the literature and its simple linear programming formulation is easy to solve. 

%In particular, we reformulate the model (\ref{Model: Wand.Parkan.minimax.disparity.LP}) in terms of the coefficients $\alpha_j$.  The coefficients alpha, obtained from our model, will be used to calculate the  OWA weights. Next, we compare our OWA weights against the OWA weights in original problem for the small dimension $n=5$ and large dimension $n=10$. Possible outcomes of this research will be:
%
%\begin{itemize}
%\item We can transform the optimization of OWA weights distribution in the optimization of coefficients alpha distribution.
%\item We can use less coefficients alpha in our proposed model in order to overcome the computational complexity when dealing with large-scale optimization problems. 
%\end{itemize}
%

We now transform the minimax disparity model (\ref{Model: Wand.Parkan.minimax.disparity.LP}) 
into a problem in which the weights are substituted by a set of coefficients
$\alpha_j, j=1, \ldots,n$,
%Solving this optimization problem, we first retrieve the coefficients $\alpha_j$  and from them we reconstruct the OWA weights.

%. In this framework, the weights of the OWA function are represented in terms of  coefficients $\alpha_j$. 
%Therefore, the objective function and all constraints in our problem (\ref{Model: our proposed model}) are written in the form of coefficients $\alpha_j$. 
%Solving the optimization problem, we obtain the optimal $\alpha_j$. Then we use $\alpha_j$ to calculate the optimal $w_i$ as their relationship shown in (\ref{OWA.binomial.decomposition.linear.system.reduced}). 

%
%\begin{eqnarray}\label{Model: our proposed model}
%\text{Min.} && \delta  \\
%\mbox{s.t.} \;  &&\sum_{i=1}^{n} \alpha_j =1 \label{eq: wi boundary}\nonumber,  \hspace{0.5cm}
%\text{Orness}(\w) = \eta = \sum_{j=1}^n \frac{n-j}{(n-1)(j+1)}\cdot\alpha_j, \label{eq: predefined level of orness}\nonumber\\
%                %
%                &&\sum_{j=1}^{n-i+1}w_{ji}\alpha_j - \sum_{j=1}^{n-i}w_{j,i+1}\alpha_j - \delta\leq 0 \label{eq: max 1} \nonumber ,  
%                \sum_{j=1}^{n-i+1}w_{ji}\alpha_j - \sum_{j=1}^{n-i}w_{j,i+1}\alpha_j + \delta\geq 0,\text{ where }
% i=1,\dots,n-1,  \label{eq: max 2}\nonumber\\
% 				%%
%				%BM condition                 
%                &&\sum_{j=2}^n \Big[1-n\frac{\binom{i-1}{j-1}}{\binom{n}{j}}\,
%\Big]\alpha_j\leq 1 \qquad \qquad \qquad \qquad
% i=2,\dots,n \nonumber
% 				%dkien wi >= 0: ko can
% 				%&& \sum_{j=1}^{n-i+1}  w_{ji}\alpha_j \geq 0
%\end{eqnarray}

\begin{eqnarray}\label{Model: our proposed model}
\text{Min.} && \delta  \\
\mbox{s.t.} \;  &&\sum_{i=1}^{n} \alpha_j =1 \label{eq: wi boundary}\nonumber,  \hspace{0.5cm}
\text{Orness}(\w) = \eta = \sum_{j=1}^n \frac{n-j}{(n-1)(j+1)}\cdot\alpha_j, \label{eq: predefined level of orness}\nonumber\\
                &&\sum_{j=1}^{n-i+1}w_{ji}\alpha_j - \sum_{j=1}^{n-i}w_{j,i+1}\alpha_j - \delta\leq 0 \qquad \qquad
 i=1,\dots,n-1, \label{eq: max 1} \nonumber\\   
                &&\sum_{j=1}^{n-i+1}w_{ji}\alpha_j - \sum_{j=1}^{n-i}w_{j,i+1}\alpha_j + \delta\geq 0 \qquad \qquad
 i=1,\dots,n-1,  \label{eq: max 2}\nonumber\\
 				%%
				%BM condition                 
                &&\sum_{j=2}^n \Big[1-n\frac{\binom{i-1}{j-1}}{\binom{n}{j}}\,
\Big]\alpha_j\leq 1 \qquad \qquad \qquad \qquad
 i=2,\dots,n \nonumber
 				%dkien wi >= 0: ko can
 				%&& \sum_{j=1}^{n-i+1}  w_{ji}\alpha_j \geq 0
\end{eqnarray}
where $0\leq \eta \leq 1$ stands for the orness of the weighting vector. Notice that the first and the last constraints in our model (\ref{Model: our proposed model}) correspond to the boundary and the monotonicity conditions of the OWA weighting vector $\w=(w_1,..., w_n) \in [0,1]^n$ with $\sum_{i=1}^n w_i=1$.

\begin{figure}[h]
\sidecaption
% For example, with the graphicx style use
\includegraphics[scale=.5]{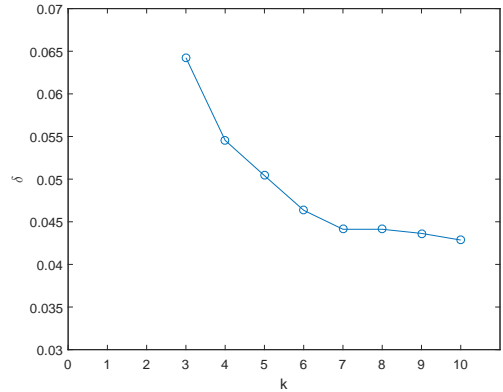}
%
% If no graphics program available, insert a blank space i.e. use
%\picplace{5cm}{2cm} % Give the correct figure height and width in cm
%
\caption{The objective value $\delta$ corresponding to the k-additive levels (for the cases $k=1, 2$ there is no solution of coefficients alpha).}
\label{fig:1}       % Give a unique label
\end{figure}

In the following we report the empirical results  with respect to  dimension $n=10$ and k-additive level is equal to $n$ (Table  \ref{Table The OWA weights generated by our proposed method using coefficients alpha for small dimension $n=10$}). We notice that a number of coefficients alpha are used is  significantly reduced in our model for the central orness values $0.3, 0.4, 0.5, 0.6$ and  $0.7$. In these cases, two non-zero coefficients  $\alpha_1$ and $\alpha_2$ are able to reconstruct  the full-dimension set of OWA weights while the remaining coefficients are zero. 
%In other words, 
%a small set of coefficients alpha can model a the full-dimensional set of OWA weights. 
%%%%%%%%%%%%%%%%%%%%%%%%%%%%%%%%
In the remaining cases of orness values, a larger number of coefficients alpha is required to generate the OWA weights. As an example, we consider our proposed model  with  the orness value
 equal to $0.2$. 
 If the k-additive level increases from $3$ to $10$,  we obtain better objective values as expected (see  Fig. \ref{fig:1}). However it is evident that $k=7$ leads to the best trade-off between the accuracy of the optimal value and the dimensionality
reduction of the optimization problem.

\begin{table}
\caption{The OWA weights generated by our proposed method  for  $n=10$}
\label{Table The OWA weights generated by our proposed method using coefficients alpha for small dimension $n=10$}      % Give a unique label
%
% Follow this input for your own table layout
%
\begin{tabular}{p{2cm}p{0.8cm}p{0.8cm}p{0.8cm}p{0.8cm}p{0.8cm}p{0.8cm}p{0.8cm}p{0.8cm}p{0.8cm}p{0.8cm}p{0.2cm}} %12 column
\hline\noalign{\smallskip}
\textit{Orness}(\w)     & {$1$} & {$0.9$} & {$0.8$} & {$0.7$} & {$0.6$}& {$0.5$} & {$0.4$} & {$0.3$} & {$0.2$}& {$0.1$} & {$0$}    \\
\hline\noalign{\smallskip}
\textit{Coefficients $\boldsymbol{\alpha}$}\\
%\noalign{\smallskip}\svhline\noalign{\smallskip}
\hline\noalign{\smallskip}
%Translation & mRNA$^a$  & 22 (19--25) & Translation repression, mRNA cleavage\\
$\alpha_1$       &10     	& 4.3	&2.71	&1.98	&1.49  	  	& 1  &0.51	&0.02	&0	&0  	& 0  \\ 
  $\alpha_2$       & -45   	& -5.4		&-1.93	&-0.98	&-0.49  		& 0  &0.49	&0.98	&0	&0	& 0 \\ 
  $\alpha_3$       & 120   	& 0			&0		&0		&0  		& 0  	&0	&0	&0	&0	& 0 \\ 
  $\alpha_4$       & -210  	& 0	&0		&0		&0  		& 0  &0	&0	&3	&0	& 0 \\ 
  $\alpha_5$       & 252  	& 12.6   	&0		&0		&0			& 0  &0	&0	&0	&0	& 0 \\ 
  $\alpha_6$       & -210  	& -16.8   	&0		&0		&0			& 0  &0	&0	&-9	&0	& 0 \\ 
  $\alpha_7$       & 120  	& 2.4   	&0		&0		&0			& 0  &0	&0	&13.71	&8.4	& 0 \\ 
  $\alpha_8$       & -45  	& 9   	&1.29		&0		&0			& 0  &0	&0	&-9.64	&-13.5	& 0 \\ 
  $\alpha_9$       & 10  	& -6.5   	&-1.57		&0		&0			& 0  &0	&0	&3.43	&7.5	& 0 \\ 
  $\alpha_{10}$       & -1  	& 1.4   	&0.5		&0		&0			& 0  &0	&0	&-0.5	&-1.4	& 1 \\
%\noalign{\smallskip}\svhline\noalign{\smallskip}
\hline\noalign{\smallskip}
\textit{OWA weights}  $\w$\\
\hline\noalign{\smallskip}
$w_1$       & 0   & 0    		&0		&0		&.05	& 0.10  &0.15	&0.20	&0.27	&0.43  & 1  \\ 
  $w_2$       & 0   & 0    		&0		&0.03		&0.06	& 0.10  &0.14	&0.18	&0.23	&0.31  & 0 \\ 
  $w_3$       & 0   & 0  	&0		&0.05		&0.07	& 0.10  &0.13	&0.16	&0.19	&0.20  & 0 \\ 
  $w_4$       & 0   & 0.01  	&0.07		&0.08		&0.10	& 0.10   &0.13	&0.1	&0.07	&0  & 0 \\ 
  $w_5$       & 0   & 0  	&0.06		&0.09		&0.10	& 0.10  &0.11	&0.11	&0.10	&0  & 0 \\ 
  $w_6$       & 0   & 0  	&0.1		&0.11		&0.11	& 0.10  &0.10	&0.09	&0.06	&0  & 0 \\ 
  $w_7$       & 0   & 0.07  	&0.14		&0.13		&0.12	& 0.10  &0.08	&0.07	&0.01	&0  & 0 \\ 
  $w_8$       & 0   & 0.20  	&0.19		&0.16		&0.13	& 0.10  &0.07	&0.05	&0	&0  & 0 \\ 
  $w_9$       & 0   & 0.31  	&0.23		&0.18		&0.14	& 0.10  &0.06	&0.02	&0	&0  & 0 \\ 
  $w_{10}$       & 1   & 0.43  	&0.27		&0.20		&0.15	& 0.10  &0.05	&0	&0	&0  & 0 \\
%\noalign{\smallskip}\svhline\noalign{\smallskip}
\hline\noalign{\smallskip}
\textbf{$\delta$} & 1 	& 0.12	&0.04	&0.02	&0.01   &0  &0.01	&0.02	&0.04	&0.12		 &1  \\
\noalign{\smallskip}\hline
\noalign{\smallskip}
\end{tabular}
%$^a$ Table foot note (with superscript)
\end{table}

%%%%%%%%%%%%%%%%%%%%%%%%%%%%%%%%%
Moreover, in Table \ref{Table The OWA weights generated by our proposed method using coefficients alpha for small dimension $n=10$}, we observe that the coefficients alpha associated with orness value $0,0.5$ and $1$ are respectively:
\begin{itemize}
\item  $\boldsymbol{\alpha}=(0,0,\ldots,1)$, which is  associated with the OWA weighting vector $\w_*=(1,0,\ldots,0)$;
\item $\boldsymbol{\alpha}=(1,0,\ldots,0)$, which is  associated with the OWA weighting vector  $\w_A=(\frac{1}{n},\ldots,\frac{1}{n})$;
\item $\boldsymbol{\alpha}=\left(\binom{n}{1},-\binom{n}{2},\binom{n}{3},\ldots,(-1)^{i-1}\binom{n}{i},\ldots, (-1)^{n-1}\binom{n}{n}\right)$, which is  associated with the OWA weighting vector $\w^*=(0,\ldots,0,1)$.
\end{itemize}

\section{Conclusions} \label{sec:5}

This study proposes  a new methodology for determining OWA weights in  large-scale optimization problems where the cost of computation of optimal weights is very high. Our  model allows  the optimization of the OWA
weights to be transformed into the optimization of  the  coefficients in the binomial decomposition framework, considering
the k-additive levels in order to reduce the complexity of the proposed model. 
The empirical result shows that a small set of the coefficients in the binomial decomposition
can model a the full-dimensional set of the OWA weights.

However, there are still some issues that need to be addressed in future research. For instance, it is necessary to develop an algorithm to identify which k-additive level in the set $\{1,\ldots, n\}$ gives the best trade off between accuracy and computational complexity according to the specific applications. 
It would also be interesting to perform a sensitivity analysis of the coefficients in the binomial decomposition with respect to the OWA weights.

%Moreover, this study challenges us with some open research questions: How can we decide a certain level $k$ in the set $\{1,\ldots, n\}$ which is considered ``good'' enough for modeling the OWA weights in certain applications? What is the sensitive analysis between OWA weights and coefficients alpha? Further empirical tests need to be carried out in higher dimensions.

\hfill\\
\noindent
\textbf{Acknowledgements} \hspace{0.1cm} The author would like to thank  Ricardo Alberto Marques Pereira and  Silvia Bortot for their helpful remarks on the manuscript.

%%%%%%%%%%%%%%%%%%%%%%%% referenc.tex %%%%%%%%%%%%%%%%%%%%%%%%%%%%%%
% sample references
% %
% Use this file as a template for your own input.
%
%%%%%%%%%%%%%%%%%%%%%%%% Springer-Verlag %%%%%%%%%%%%%%%%%%%%%%%%%%
%
% BibTeX users please use
% \bibliographystyle{}
% \bibliography{}
%
%\bibliographystyle{spbasic}

%current used this 
\bibliographystyle{spmpsci}
%\bibliography{DOW-bibliografia} % full name journal
\bibliography{DOW_bibliografia_abbreviation} % abbreviation journal name

\end{document}